\documentclass[11pt]{article}

\usepackage[margin=1in]{geometry}
\usepackage{amsmath}        % math environments like align, aligned, etc.
\usepackage{amsfonts}       % blackboard math symbols
\usepackage{bm}             % bold math symbols
\usepackage{booktabs}       % professional-quality tables
\usepackage{graphicx}       % graphics
\usepackage{subcaption}     % subfigure environment
\usepackage{url}            % simple URL typesetting
\usepackage{textcomp}
\usepackage[numbers,sort&compress]{natbib}
\usepackage[hidelinks]{hyperref} % hyperlinks (load last)

\begin{document}

\title{Neural Controlled Differential Equations for EMT-Level Surrogate Modeling of Grid-Forming Inverters}

\author{
Jiagang Qu,
Yong Tao,
Dan Wang,
Enyi Li,
Jingjing Qi,
and Ding Wang\thanks{Corresponding author: \texttt{ding2.wang@midea.com}.}\\
\small Corporate Research Center, Midea Group, China
}

\date{}

\maketitle

\begin{abstract}
The application of artificial intelligence methods in power electronic converter modeling is becoming increasingly widespread, but existing applications still face many challenges, such as difficulties in multi-time-scale hybrid analysis and the lack of physics-aware evaluation criteria and constraints, resulting in poor performance. This paper proposes a Neural Controlled Differential Equation (Neural CDE) framework for learning continuous-time surrogate models of grid-forming inverters for electromagnetic transient (EMT) simulation, which relaxes the constraint of fixed sampling rates and enables multi-time-scale control analysis. Then, an affine-control formulation with dual slow/fast pathways is proposed to capture the hierarchical and multiscale behavior of converter dynamics, and a physics-inspired regularization method is utilized to enhance stability and coherence. Evaluated on EMT-generated trajectories, the model accurately reproduces transient responses, preserves effective damping and the dominant oscillatory characteristics, and maintains bounded long-horizon rollouts. The results show that Neural CDE-based component modeling offers a physically consistent surrogate modeling approach for EMT-level simulation studies.
\end{abstract}


\section{Introduction}
With the explosive growth of renewable energy resources and energy storage systems, grid-forming converters are widely used to mimic the behavior of synchronous generators (SG) and thereby increase grid strength \cite{altalaq2025asophisticated, lei2024transient, huang2019transient}. However, the individual capacity of synchronous generators typically reaches several hundred megawatts \cite{kundur1994power}, whereas grid-forming converters are usually only a few hundred kilowatts. Therefore, it takes thousands of grid-forming converters to form a station comparable to a single synchronous machine, leading to a dramatic increase in power system complexity. To analyze the multi-machine interaction characteristics of such an energy-storage station, traditional electromagnetic transient (EMT) simulation methods for converters are slow and typically limited to simulating systems composed of dozens of devices at most, making them incapable of analyzing complex systems with over 1,000 devices. Additionally, electromagnetic transient models require all parameters to be known, yet equipment manufacturers usually do not disclose their detailed control methods and parameters \cite{li2021impedance}. Therefore, current electromagnetic transient simulation methods cannot accommodate the mixed application of devices from different manufacturers, motivating a black-box and physically consistent surrogate modeling approach.

Artificial intelligence-based methods enable the construction of black-box models directly from measurement or simulation data, allowing the complex behavior of power electronic converters to be captured with high fidelity. Owing to their strong nonlinear approximation capability, these data-driven models remain effective even for highly complex systems, making neural networks an attractive tool for both steady-state and transient-state modeling of inverters in large-scale power electronic systems.

For steady-state problems such as power flow analysis, Graph Neural Networks (GNNs) have been widely employed to approximate the mapping between network topology, operating conditions, and nodal voltages, thereby enabling fast and scalable solutions \cite{donon2019graph, hamann2024foundation}. In contrast, the transient behavior of power systems is considerably more complex, characterized by nonlinear, stiff, and high-dimensional dynamics. Existing neural approaches can be broadly categorized into two paradigms: (1) data-driven models, which directly learn dynamic input-output relationships using architectures such as Recurrent Neural Networks (RNNs) \cite{qashqai2020modeling}, and (2) physics-informed models \cite{nellikkath2024physics}, which embed physical priors or constraints into neural formulations. However, these approaches often struggle to capture the intricate, nonlinear dynamics of large-scale power systems. A more general and flexible paradigm is to model complex electronic components, such as inverters, using neural networks, and then embed these learned component models within traditional numerical simulation frameworks \cite{li2022neural, xiao2022feasibility, moya2023approximating, guruwacharya2024data, shen2025neuro}. This allows leveraging the efficiency of neural surrogates while preserving the physical interpretability and stability of conventional iterative solvers.

Neural differential equation models provide a principled way to learn continuous-time component dynamics and integrate them into conventional simulation loops. In particular, Neural ODEs \cite{chen2018neural} have been explored as continuous-time surrogates for power-system and power-electronic component dynamics. For example, Li \emph{et al.} \cite{li2022neural} integrate Neural ODE-based component models into EMT simulation loops to improve scalability, and Xiao \emph{et al.} \cite{xiao2022feasibility} employ Neural ODE surrogates for inverter dynamics in transient simulation studies. Despite their promise, directly adopting Neural ODEs in inverter-dominated EMT settings still leaves key challenges insufficiently addressed.

First, many learning-based transient models are tightly coupled to specific sampling rates and solver settings, which can degrade generalization when the EMT time step or control update rate changes. Second, control inputs in practical grid-forming inverters are inherently structured and multi-rate: fast inner-loop references and modulation-related signals coexist with slower supervisory commands. Treating inputs as piecewise-constant or purely exogenous signals obscures this hierarchy and makes it difficult to faithfully capture multiscale closed-loop behavior. Third, unconstrained neural dynamics embedded in closed-loop EMT simulation may exhibit spurious oscillations, energy drift, or unstable long-horizon rollouts, especially under stiff EMT-level time steps and large-scale interconnections. Moreover, without physics-aware evaluation criteria, it is challenging to verify whether learned surrogates preserve key transient properties.

These considerations motivate a continuous-time, control-aware, and stability-conscious modeling framework tailored to grid-forming inverter dynamics for large-scale EMT simulation. Neural Controlled Differential Equations (Neural CDEs) extend Neural ODEs by explicitly parameterizing how latent states are driven by a control path, making them well suited to inverter systems where discretely sampled, multi-rate control signals play a central role. By representing control trajectories as continuous paths and learning the state-dependent response to these paths, Neural CDEs offer a natural mechanism to capture hierarchical control effects while remaining consistent with EMT continuous-time dynamics.

The key contributions of this work are summarized as follows:
\begin{enumerate}
    \item \emph{Unified Neural CDE framework.} A Neural CDE formulation is developed for modeling grid-forming inverter dynamics under symmetric operation, enabling continuous-time learning from discretely sampled control signals.
    \item \emph{Affine-control decomposition.} The proposed affine form separates intrinsic system dynamics and control-driven effects, aligning with the inverter's hierarchical control structure and capturing multi-time-scale behaviors.
    \item \emph{Dual-path control embedding.} A dual control pathway, representing slow and fast variations of control inputs, is introduced to enhance the model's expressiveness and stability across operating conditions.
    \item \emph{Physics-informed regularization and evaluation for EMT-level deployment.} We incorporate a Jacobian-based stability regularizer tailored to typical EMT time steps and propose physics-coherent metrics that assess resonance preservation and effective damping via envelope decay under 1~kHz sampling. These tools jointly promote bounded long-horizon rollouts and alignment with the closed-loop behavior observed in EMT simulations.
\end{enumerate}

\section{Modeling System and Fundamental Concept}
\subsection{Grid-Forming Inverter}
\begin{figure}
    \centering
    \includegraphics[width=0.85\textwidth]{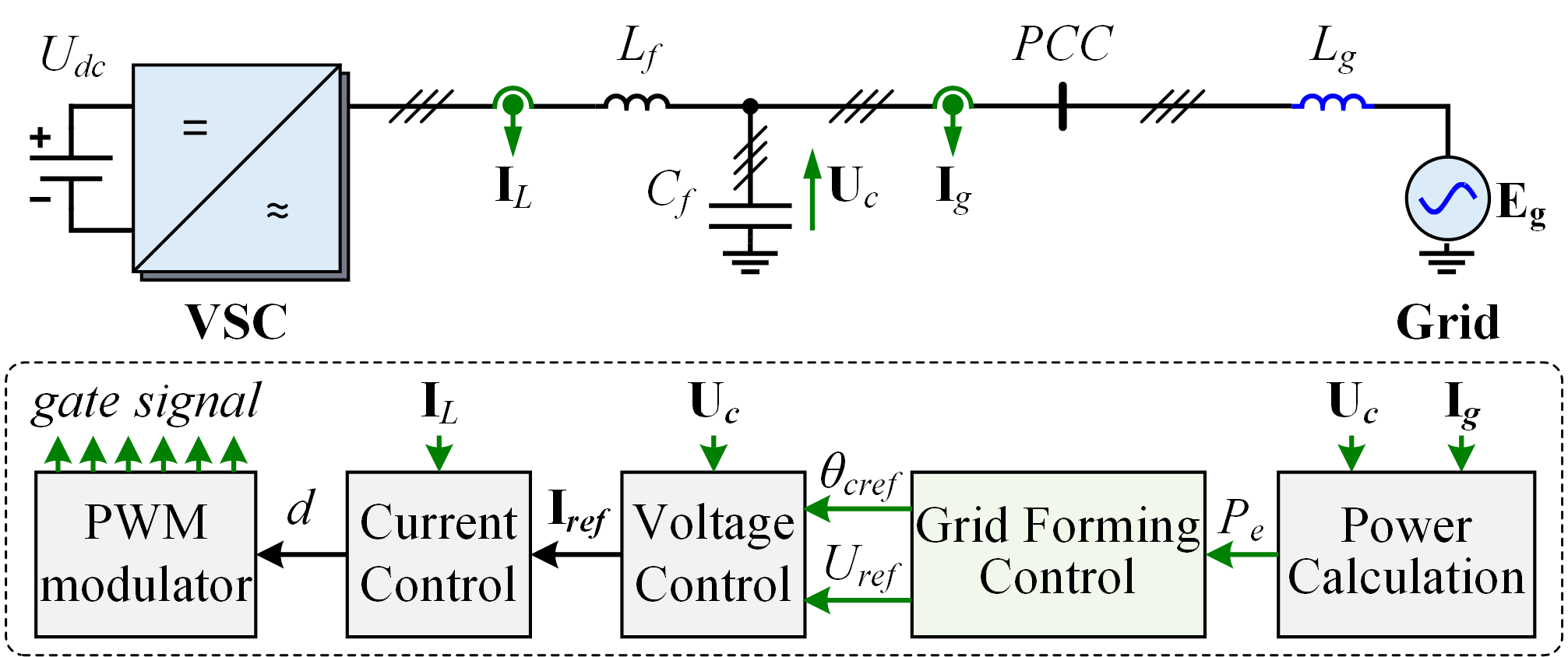}
    \caption{Structure of Grid Forming Converter.}
    \label{fig:gfm_converter}
\end{figure}

In this work, we focus on modeling the grid-forming inverter, which is crucial for the frequency regulation and voltage support of power systems. Fig.~\ref{fig:gfm_converter} demonstrates the typical control and circuit configuration of a grid-tied GFM converter. The GFM is interfaced with the point of common coupling (PCC) via a conventional LC filter network, where $\bm E_g$ and $\bm U_c$ are the grid voltage and capacitor voltage, and $\bm I_g$ and $\bm I_L$ are the grid current and inductor current, respectively. The DC bus voltage $U_{dc}$ is considered to be constant as it is typically maintained by other components like a battery storage system. The control scheme commonly employs an inner dual-loop control structure to track the PCC voltage to its target value, where the reference signal $\bm U_{ref}$ comes from the outer control loop. A typical GFM control structure with voltage feed-forward control is considered in this paper, which avoids control saturation in the inner loop.

Then, the dynamics of a grid-forming inverter can be described by a set of differential equations that capture the behavior of its internal components, such as the voltage source converter (VSC), filters, and control systems.
\begin{equation}
    \frac{d\bm x}{dt} = f(\bm x, \bm u)
\end{equation}
where $\bm x$ represents the state variables of the inverter and $\bm u$ denotes the control inputs, and $f$ is a nonlinear function that describes the system dynamics.

The goal of the neural network is to learn a mapping $f_\theta$ that can accurately approximate the derivatives of the state variables given the current state and control inputs.
This learned model can then be integrated into a traditional simulation framework to predict the inverter's behavior over time.
This process can be summarized as follows:
\begin{align}
    \frac{d\bm x(t)}{dt} &\approx f_\theta(\bm x(t), \bm u(t)) \\
    \bm x(t) &= \bm x(t_0) + \int_{t_0}^{t} f_\theta(\bm x(t), \bm u(t)) dt
\end{align}
where $\bm x(t), \bm u(t)$ is the state and controlled vector at time $t$, and $\bm x(t_0)$ is the initial state at time $t_0$.
In this work, the state vector is the $dq0$ grid current $\bm i_{dq0}$, and the control vector comprises the grid-forming controller reference inputs---the commanded voltage magnitude $A$ and phase $\phi$---together with the $dq0$ components of the measured LC filter capacitor voltage $\bm v_{dq0}$.

Once the neural network is trained, it can be used to simulate the inverter's dynamics by integrating the learned differential equations over time using numerical solvers, and the model can be embedded within larger power system simulations to analyze the overall system behavior.

\subsection{Neural ODEs and Neural CDEs}
To capture continuous-time EMT dynamics while remaining compatible with numerical simulation loops, we adopt neural differential equation models.

Neural ODEs \cite{chen2018neural} model the derivatives of system states with a neural network,
\begin{equation}
    \dot{\bm{x}}(t)=f_\theta(\bm{x}(t), t),
\end{equation}
and recover trajectories by integrating with an ODE solver. For forced or controlled systems, one commonly augments the dynamics with an input signal $\bm{u}(t)$,
\begin{equation}
    \dot{\bm{x}}(t)=f_\theta(\bm{x}(t), \bm{u}(t), t),
\end{equation}
or adopts structured neural ODE variants such as spline/controlled neural ODEs \cite{quaglino2019snode}. A key appeal is solver compatibility: the learned dynamics can be embedded into the standard time-stepping frameworks used in EMT simulation.

However, in grid-forming inverters, control signals are typically available only at discrete update instants (often multi-rate), whereas EMT solvers advance at a possibly different time step. Directly treating $\bm{u}(t)$ as continuous or piecewise-constant may obscure the control hierarchy and can introduce sensitivity to sampling and solver settings. Neural CDEs \cite{kidger2020neural}, grounded in controlled differential equations \cite{lyons2007differential}, address this mismatch by driving the latent dynamics with a continuously interpolated control path $\bm{X}(t)$ constructed from discrete observations:
\begin{equation}
    \mathrm{d}\bm{x}(t)=f_\theta(\bm{x}(t))\,\mathrm{d}\bm{X}(t).
\end{equation}
By explicitly modeling how the state evolves with respect to the control path, Neural CDEs naturally handle irregular sampling and provide a principled interface between discretely updated controls and continuous-time EMT dynamics.

\section{Methodology}

Neural CDEs are adapted to model the inverter dynamics under discrete control inputs.
The dynamics are defined as follows:
\begin{align}
    d\bm x(t) &= f_\theta(\bm x(t)) dX(t) \\
    \bm x(t) &= \bm x(t_0) + \int_{t_0}^{t} f_\theta(\bm x(t)) dX(t)
\end{align}
where $X(t)$ denotes a continuous path constructed from the discrete observations $\{(u_i, t_i)\}_{i=0}^N$ using interpolation methods such as linear or cubic splines.

For power electronic systems, it is often convenient to make explicit the separation between autonomous dynamics and control-driven effects.
To this end, we adopt an affine-control Neural CDE of the form
\begin{align}
    \label{neural-cde-affine}
    d\bm x(t) &= f_\theta(\bm x(t)) dt + g_\theta(\bm x(t)) dU(t)
\end{align}
where $U(t)$ represents the interpolated control path without time $\{(u_i)\}_{i=0}^N$, and $g_\theta$ captures the control-dependent dynamics modulated by control inputs.
This affine formulation can be parameterized as a standard Neural CDE but allows the model to separately learn the effects of control inputs and inherent system dynamics, thereby enhancing its generalization capability across operating scenarios.

To further enrich the model's expressiveness, we draw inspiration from Neural ODEs with controlled inputs and perform a first-order Taylor expansion of the control-dependent term,
\[
f_\theta(\bm x(t), \bm u(t))
= f_\theta(\bm x(t), 0)
+ \left.\frac{\partial f_\theta}{\partial \bm u}\right|_{\bm u=0} \bm u(t)
+ O(\|\bm u(t)\|^2).
\]
Neglecting higher-order terms leads to the following first-order approximation:
\begin{equation}
\label{first-order-ode}
f_\theta(\bm x(t), \bm u(t))
\approx f_\theta(\bm x(t)) + g_\theta(\bm x(t)) \bm u(t),
\end{equation}
where $f_\theta(\bm x(t),0)$ represents the autonomous dynamics and
$g_\theta(\bm x(t)) := \left.\partial f_\theta / \partial \bm u \right|_{\bm u=0}$
captures the control-dependent effects.
This shows that, by choosing the control path $U(t)$ such that $dU(t) = \bm u(t) dt$, the Neural CDE formulation can represent the standard first-order controlled ODE structure.
In many Neural CDE applications, the driving path $X(t)$ is simply taken to be the observed input signal (or even the observed state trajectory itself), so that the model effectively conditions on the historical evolution of the system.

In this work, we incorporate both types of control inputs together with an explicit residual term to enhance the representational capacity of the model. The resulting Neural CDE is formulated as
\begin{equation}
    \label{neural-cde-dynamics}
    \begin{aligned}
        d\bm x(t) &= f_\theta(\bm x(t))\, dt + \sum_{i=1}^2 g_\theta^{(i)}(\bm x(t))\, dX_i(t) \\
                  &\quad + h_\theta(\bm x(t), \bm u(t))\, dt ,
    \end{aligned}
\end{equation}
where the control paths are defined by
\begin{equation}
\label{neural-cde-controls}
    dX_1(t) = \bm u(t)\, dt, \qquad
    dX_2(t) = d\bm u(t).
\end{equation}
Here, $X_1(t)$ captures the slow variations of the control commands, while $X_2(t)$ represents their rapid changes. The additional drift term $h_\theta(\bm x(t), \bm u(t))$ is introduced to model the residual effects of higher-order terms neglected in the truncated Taylor expansion.

This combined control input strategy enables the model to exploit both the instantaneous control commands and their rapid temporal variations, which is critical for capturing the fast and nonlinear switching dynamics of grid-forming inverters, and ultimately improves generalization across operating conditions.

% augmentation technique
A known limitation of standard Neural ODEs is that their flow maps must be diffeomorphisms; that is, the learned dynamics cannot fold or merge distinct regions of the state space. Augmented Neural ODEs \cite{dupont2019augmented} overcome this limitation by lifting the dynamics into a higher-dimensional latent space, where the flow becomes invertible, and then projecting back to the original state space. The same principle applies to Neural CDEs, yielding Augmented Neural CDEs that combine continuous-time controlled dynamics with enhanced representational capacity.

\section{\label{sec-theory} Theoretical Insights and Analysis}

Neural CDEs offer a continuous-time representation that is compatible with nonlinear input--state models and reflects the dissipative behavior of power electronic systems.
In the affine-control form, the model separates slow outer-loop effects and fast inner-loop variations, matching the hierarchical control structure of grid-forming inverters.

\subsection{Controllability and Multi-Timescale Structure}

The affine Neural CDE in \eqref{neural-cde-dynamics} naturally aligns with the hierarchical control structure of grid-forming inverters. The slow path \(X_1(t)\) represents supervisory control actions, such as droop and Virtual Synchronous Machine (VSM) mechanisms and outer voltage--power regulation, that evolve on a low bandwidth and determine the quasi-steady operating point of the converter.

The fast path \(X_2(t)\) captures the high-bandwidth electromagnetic dynamics shaped by the inner current-control loops.
These dynamics operate at sub-millisecond scales and dominate the rapid evolution of inductor currents and capacitor voltages.

By separating these effects into slow and fast input channels, the Neural CDE formulation mirrors the intrinsic multi-time-scale structure of inverter control.
This decomposition preserves the controllability interpretation of nonlinear input--state systems and mitigates numerical stiffness when the learned dynamics are embedded in EMT-level simulation.

\subsection{Numerical Stability via Jacobian Regularization}

To improve the numerical behavior of the learned dynamics when embedded into electromagnetic transient (EMT) solvers, we regularize the Jacobian of the discrete-time flow induced by the neural vector field, inspired by~\cite{finlay2020train}.

Let $\bar f_\theta(x)$ denote the learned autonomous dynamics and $J_\theta(x) = \partial_x \bar f_\theta(x)$ its Jacobian.
For a given nominal EMT time step $\Delta t$, the one-step integration defines a discrete-time flow map $\Phi_\theta^{\Delta t}(x)$ with Jacobian
$D_x \Phi_\theta^{\Delta t}(x)$.
For small $\Delta t$, this Jacobian admits the first-order approximation
\begin{equation}
    D_x \Phi_\theta^{\Delta t}(x)
    = I + \Delta t\, J_\theta(x) + \mathcal{O}(\Delta t^2).
\end{equation}
We therefore use
\begin{equation}
    A_\theta(x) \approx I + \Delta t\, J_\theta(x)
\end{equation}
as a proxy for the discrete-time Jacobian and penalize spectral radii outside the unit disk:
\begin{equation}
    \mathcal{L}_{\mathrm{stab}}
    =
    \mathbb{E}\Big[
        \max\big(0,\,
        \rho(A_\theta(x)) - 1
        \big)
    \Big].
\end{equation}
Although training employs a fixed-step Runge--Kutta (RK4) integrator and inference uses a higher-order adaptive solver (dopri5), this regularizer controls the local sensitivity of the discrete-time flow for typical EMT step sizes and is intended to suppress unstable modes and long-horizon divergence.

\subsection{\label{sec-theory-physics} Physics-Coherent Dynamics under Partial Observability}

In this paper, only the $dq0$ grid current $\bm i_{dq0}$ and the LC filter capacitor voltage $\bm v_{dq0}$ are available, sampled at 1~kHz.
Under this limited sampling bandwidth, the kilohertz-range resonance of the physical LC network is strongly suppressed by the closed-loop current controller, PWM modulation, and the EMT discretization scheme.
As a result, the measured trajectories primarily exhibit low-frequency closed-loop dynamics rather than the bare LC resonance.

The Neural CDE model is trained directly on these partially observable trajectories and thus learns an effective continuous-time vector field consistent with the sampled EMT behavior.
To assess physics coherence under these constraints, we qualitatively examine the response of $i_g$ to step-like perturbations.
Across operating points, the Neural CDE model reproduces bounded trajectories and recovers the same qualitative pattern of decaying oscillations observed in the EMT simulations, indicating that the surrogate preserves the essential closed-loop behavior required for EMT-level studies.

To further quantify the agreement in effective damping, we extract a continuous-time amplitude envelope using the analytic signal.
Let
\begin{equation}
    z(t) = i_g(t) + j \mathcal{H}[i_g(t)]
\end{equation}
denote the Hilbert transform-based analytic signal, and extract the amplitude envelope $A(t) = \vert z(t) \vert$.
Assuming an exponentially decaying envelope $A(t) \approx A_0 e^{-\sigma t}$, the damping rate $\sigma$ is obtained by a least-squares fit of $\log A(t)$ to a straight line.
A normalized damping-score is then defined as
\begin{equation}
S_{\text{damp}}
= 1 - \bigg| \frac{\sigma_{\text{CDE}} - \sigma_{\text{EMT}}}{\sigma_{\text{EMT}}} \bigg|,
\end{equation}
which equals one when the Neural CDE model matches the EMT decay rate and decreases as the discrepancy grows. The detailed results are presented in Section~\ref{sec-exp-physical-consistency}.

At the same time, precisely matching the effective damping rate from low-rate, noisy envelope estimates proves challenging.
This limitation reflects both the partial observability and the sampling bandwidth, and points to an interesting direction for future work on more tightly constrained, physics-aware training objectives.

\section{Experiments}

In this section, an ablation study and a model sensitivity analysis are conducted to assess the impact of different components of the proposed methodology.

\subsection{\label{implementation} Implementation and Training Setup}

Neural ODE-based models generally benefit from smooth and well-conditioned input trajectories, which improve both training stability and generalization capability.
To this end, a low-pass filter is applied to the raw simulation data to suppress high-frequency noise and transient components that may otherwise impede the learning of the underlying physical dynamics.

After filtering, all signals are normalized in per-unit form to ensure that input features and state variables are on a comparable numerical scale, thereby facilitating efficient optimization.
The reference base values are set to 245~A for current and 1000~V for voltage.

To determine an appropriate integration time step for the ODE solver, the dominant dynamic time scales of the training trajectories are analyzed using the Autocorrelation Function (ACF), Power Spectral Density (PSD), and Accumulated Energy Percentage (AEP) methods.
This analysis ensures that the solver captures essential dynamics while avoiding unnecessary computational overhead.

During training, the Runge-Kutta 4 (RK4) method is employed as the ODE solver owing to its balance between stability and computational efficiency.
For inference, the adaptive Dormand-Prince 5(4) (DOPRI5) method is used to achieve higher precision with adaptive step size control.

The proposed Neural CDE framework comprises four feedforward subnetworks: $f_\theta$, $g_\theta^{(1)}$, $g_\theta^{(2)}$, and $h_\theta$.
The networks $f_\theta$, $g_\theta^{(1)}$ and $g_\theta^{(2)}$ share a common trunk architecture but employ separate head layers.
The baseline trunk consists of two hidden layers with 256 neurons each, while the head layers are linear mappings whose output dimensions match the state dimension.
The residual module is implemented as a single linear layer to model higher-order correction terms.
All subnetworks use the $\tanh$ activation, and the deployed model additionally augments the state with extra latent dimensions following the Augmented Neural CDE formulation.

The models are trained with the AdamW optimizer (learning rate $10^{-3}$, weight decay $10^{-4}$) for up to 10{,}000 epochs, minimizing the mean-squared error between predicted and reference $dq0$ trajectories over short integration windows of 26 samples.
To stabilize training under stiff dynamics, gradient clipping combining adaptive gradient clipping (AGC), norm clipping, and value clipping is applied.
The complete set of training hyperparameters is summarized in Table~\ref{tab:hyperparams}.

\begin{table}
    \centering
    \caption{Training and network hyperparameters.}
    \label{tab:hyperparams}
    \begin{tabular}{lc}
        \toprule
        Hyperparameter & Value \\
        \midrule
        Optimizer               & AdamW \\
        Learning rate           & $10^{-3}$ \\
        Weight decay            & $10^{-4}$ \\
        Gradient clipping       & AGC + norm + value \\
        Activation              & $\tanh$ \\
        Trunk hidden width      & 256 \\
        Training window length  & 26 samples \\
        Training solver         & RK4 (fixed step) \\
        Inference solver        & DOPRI5 ($\mathrm{rtol}=10^{-4}$, $\mathrm{atol}=10^{-6}$) \\
        Max epochs              & 10{,}000 \\
        \bottomrule
    \end{tabular}
\end{table}

\subsection{Simulation Setup}

The target system is a grid-connected energy-storage converter built around an Active Neutral-Point-Clamped (ANPC) three-level topology, interfaced to the grid through an LCL filter.
On the DC side, the converter is fed by a stiff 2000~V bus supported by a 1~mF capacitor that represents the aggregated storage and decoupling capacitance.
On the AC side, the LCL filter consists of a converter-side inductor of 180~$\mu$H, a grid-side inductor of 36~$\mu$H, and a 40~$\mu$F filter capacitor.
In the surrogate model, the grid-side inductor is absorbed into the line inductance, so the filter is treated as an effective LC network.
The point of common coupling is connected to a balanced three-phase grid with a line-to-line RMS voltage of 1000~V at 50~Hz, through a line impedance of 30~m$\Omega$ and 0.5~mH per phase.
All EMT trajectories used for training and evaluation are generated from this configuration under symmetric operation. The complete set of electrical parameters is summarized in Table~\ref{tab:sim-setup}.

\begin{table}
    \centering
    \caption{Electrical parameters of the simulated grid-connected ANPC energy-storage converter.}
    \label{tab:sim-setup}
    \begin{tabular}{lcc}
        \toprule
        Parameter & Symbol & Value \\
        \midrule
        Converter topology          & --        & ANPC three-level \\
        DC-bus voltage              & $V_{dc}$  & 2000~V \\
        DC-bus capacitance          & $C_{dc}$  & 1~mF \\
        Converter-side inductor     & $L_1$     & 180~$\mu$H \\
        Grid-side inductor          & $L_2$     & 36~$\mu$H \\
        Filter capacitor            & $C_f$     & 40~$\mu$F \\
        Grid voltage (line-to-line RMS) & $V_g$ & 1000~V \\
        Grid frequency              & $f_g$     & 50~Hz \\
        Line resistance             & $R_l$     & 30~m$\Omega$ \\
        Line inductance             & $L_l$     & 0.5~mH \\
        \bottomrule
    \end{tabular}
\end{table}

\subsection{Dataset Description}

We evaluate the proposed Neural CDE model on a dataset generated from detailed EMT simulations of the converter described in Table~\ref{tab:sim-setup} under symmetric operation.
The dataset comprises 90 independent trajectories spanning various operating scenarios, such as load changes, grid disturbances, and reference signal variations.
Each trajectory has a duration of approximately 5.1~s.
The raw EMT waveforms are simulated at 100~kHz; after an anti-aliasing low-pass filter, the signals are decimated by a factor of 100 to a sampling rate of 1~kHz ($\Delta t = 1$~ms), yielding 5101 samples per trajectory.
The three-phase grid currents and terminal voltages are transformed into the $dq0$ frame using a power-invariant Park transformation.
Accordingly, the model state is the three-dimensional $dq0$ grid current $\bm i_{dq0}$, and the control input is a five-dimensional vector consisting of the commanded voltage magnitude $A$ and phase $\phi$ together with the $dq0$ terminal voltages $\bm v_{dq0}$.
All channels are normalized to per-unit form using the base values reported in Section~\ref{implementation}.
The 90 trajectories are randomly partitioned into 72 training, 9 validation, and 9 test cases (an 80/10/10 split) with a fixed random seed.

\begin{figure}[!t]
    \centering
    \includegraphics[width=\textwidth]{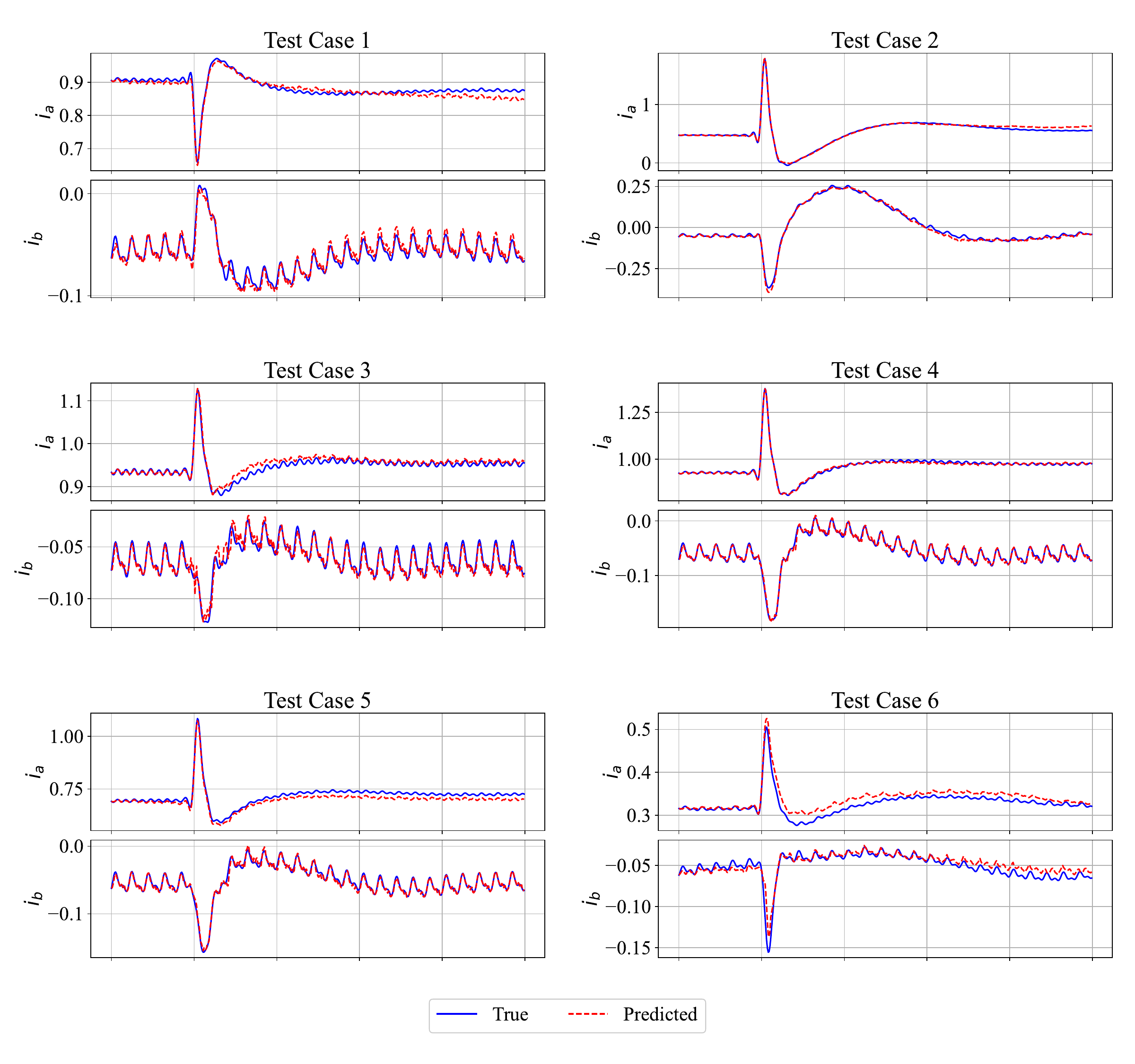}
    \caption{Comparison of grid current in the $dq$ frame between EMT simulation and the Neural CDE model under various test cases.}
    \label{fig:current_dq}
\end{figure}

\begin{figure}[!t]
    \centering
    \includegraphics[width=\textwidth]{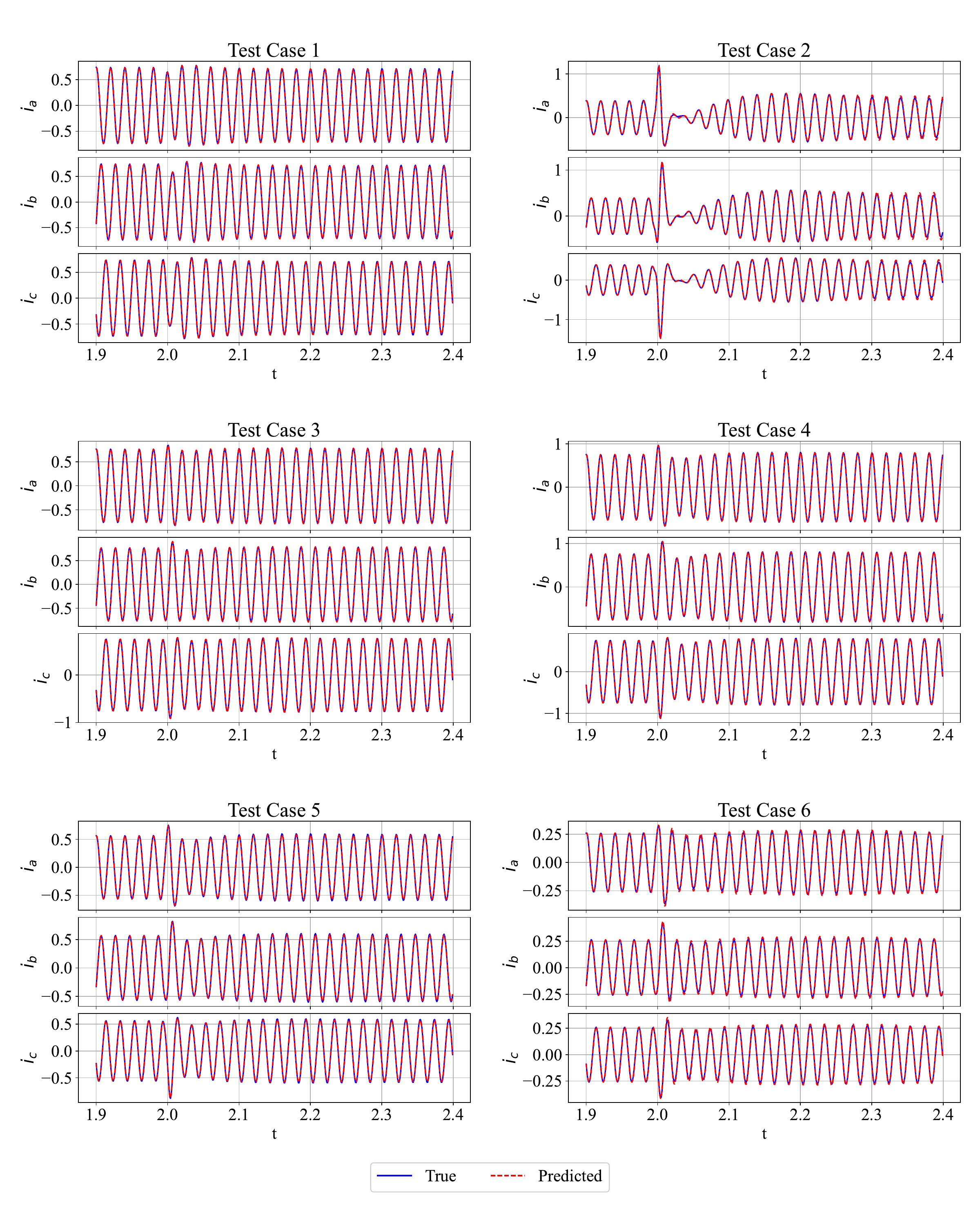}
    \caption{Comparison of grid current in the $abc$ frame between EMT simulation and the Neural CDE model under various test cases.}
    \label{fig:current_abc}
\end{figure}

\begin{figure}[!t]
    \centering
    \includegraphics[width=0.9\textwidth]{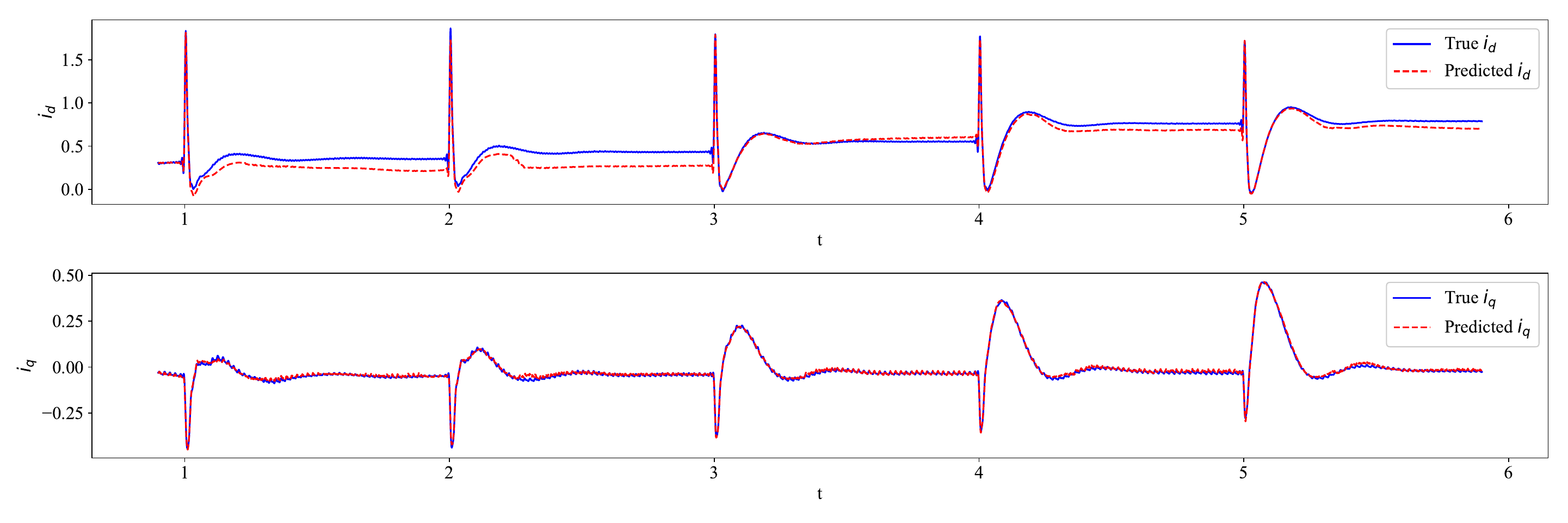}
    \caption{Long-horizon rollout of the Neural CDE model showing bounded state trajectories over extended integration time.}
    \label{fig:long_horizon}
\end{figure}

\subsection{Ablation Study}

Each variant modifies only one component of the proposed framework while keeping all others fixed.
We adopt a controlled Neural ODE, i.e. $d\bm x/dt = f_\theta(\bm x, \bm u(t))$ with the discrete control interpolated into a continuous input, as the baseline.
This baseline receives identical state and control inputs and is sized to a comparable parameter budget (about $2.0\times10^{5}$ parameters, within $3\%$ of the proposed models), so that the reported gains can be attributed to the controlled differential formulation rather than to differences in model capacity.
Model performance is evaluated in terms of prediction accuracy and trajectory stability, where the stability score quantifies how strongly the prediction error amplifies over long-horizon integration.

We quantify numerical stability using a Lyapunov-style error growth indicator.
Let $e(t) = \lVert \hat{\bm x}(t) - \bm x(t) \rVert$ denote the
instantaneous state error.
\begin{equation}
    \text{Stability Score} = 1 - \frac{1}{T-\Delta t} \int_0^{T-\Delta t} \Bigg[ \log\frac{e(t+\Delta t)+\varepsilon}{e(t)+\varepsilon} \Bigg]_+ dt
\end{equation}
The integrand retains only the positive part of the local logarithmic error-growth rate, so the score equals one when the prediction error never amplifies along the rollout and decreases as the error grows more rapidly; values approaching or below zero correspond to diverging, unstable trajectories.

\paragraph{Neural ODEs vs. Neural CDEs.}
Switching from Neural ODEs to Neural CDEs yields the largest single-component improvement.
The MSE decreases from roughly $6.0726\times10^{-5}$ to $3.8649\times10^{-5}$, and stability increases from 0.8220 to 0.8374.
Neural ODEs occasionally drift when the control input changes abruptly, whereas Neural CDEs maintain smoother and more bounded rollouts, confirming that explicitly modeling controlled dynamics better captures inverter transients.

\paragraph{Affine vs. Standard Neural CDEs.}
Introducing the affine decomposition further reduces error to $3.5407\times10^{-5}$ and slightly improves stability.
Training curves become notably smoother: the gradient noise is reduced, the loss spikes vanish, and both the training and validation losses exhibit monotonic decay over a much longer horizon.
This suggests that separating intrinsic dynamics $f_\theta$ and control-induced effects $g_\theta$ makes the vector field easier to optimize and leads to a more stable learned flow.

\paragraph{Single vs. Dual Control Pathways.}
Adding the dual fast/slow control pathways provides another substantial improvement: MSE decreases to $2.7681\times10^{-5}$, and stability reaches 0.8463, the highest among all single ablations.
This confirms that LC-type inverters exhibit strong multiscale behavior, where high-frequency switching and slow envelope variations affect the dynamics through distinct channels.

\paragraph{Effect of Residual Module $h_\theta$.}
The residual module produces smaller but consistent gains.
While the affine decomposition already captures the majority of control-driven structure, adding $h_\theta$ helps correct high-order nonlinearities and fast perturbations that are difficult to express using only $f_\theta$ and $g_\theta$.

The ablation results demonstrate that the Neural CDE formulation, affine decomposition, and dual control pathways deliver the largest benefits, while the residual block serves as a fine-grained refinement mechanism.

\begin{table}
    \centering
    \caption{Ablation study results showing the impact of different model components on parameter count, MSE, and stability score. All variants are kept within a comparable parameter budget so that the reported gains reflect the model formulation rather than capacity.}
    \label{tab:ablation}
    \footnotesize
    \setlength{\tabcolsep}{4pt}
    \begin{tabular}{lccc}
        \toprule
        Model Variant           & \# Params & MSE ($\times10^{-5}$) & Stability \\
        \midrule
        Neural ODEs             & 200{,}451 & 6.0726          & 0.8220 \\
        Standard Neural CDEs    & 203{,}026 & 3.8649          & 0.8374 \\
        Affine Neural CDEs      & 203{,}026 & 3.5407          & 0.8385 \\
        Dual Control Pathway    & 206{,}881 & 2.7681          & \textbf{0.8463} \\
        With Residual Module    & 203{,}053 & 3.0909          & 0.8440 \\
        Dual Control + Residual & 206{,}908 & \textbf{2.6853} & 0.8449 \\
        \bottomrule
    \end{tabular}
\end{table}

\subsection{Model Sensitivity Analysis}

The goal of model sensitivity analysis is to quantify the marginal performance gain obtained by expanding each module under a fixed parameter budget.

\paragraph{Trunk-Base Capacity Allocation}
Table~\ref{tab:trunk-base} compares models with identical total parameter count but different trunk/base splits.
Performance improves as more capacity is shifted toward the base layers, reaching the best configuration at in-256-256 trunk + 256-256-256 base, which attains the lowest MSE of $3.5407\times10^{-5}$ while maintaining high stability of 0.8385.
This indicates that the base dynamics contribute more to expressive power than further deepening the trunk encoder once moderate trunk depth is available.

\begin{table}
    \centering
    \caption{Model sensitivity analysis results showing the effect of trunk-base capacity allocation.}
    \label{tab:trunk-base}
    \footnotesize
    \setlength{\tabcolsep}{4pt}
    \begin{tabular}{lccc}
        \toprule
        Trunk layers       & Base layers              & MSE ($\times10^{-5}$) & Stability \\
        \midrule
        in-256-256-256-256 & 256-out                  & 4.3863          & 0.8348 \\
        in-256-256-256     & 256-256-out              & 3.6181          & \textbf{0.8402} \\
        in-256-256         & \textbf{256-256-256-out} & \textbf{3.5407} & 0.8385 \\
        in-256             & 256-256-256-256-out      & 3.7068          & 0.8386 \\
        \bottomrule
    \end{tabular}
\end{table}

\paragraph{Marginal Gain of Base, Fast, Slow, and Residual Branches}
Using the optimal trunk-base configuration from Table~\ref{tab:trunk-base} together with the best-performing ablation variant as the baseline, Table~\ref{tab:sensitivity} evaluates the effect of adding one layer to each branch:
\begin{itemize}
    \item Residual branch gives the largest MSE reduction ($2.5204\times10^{-5}$, $\approx$6\% improvement).
    \item Fast branch yields the highest stability (0.8518, the best overall).
    \item Slow branch improves accuracy modestly, consistent with its role in capturing low-frequency envelopes.
    \item Base branch yields small yet steady gains.
\end{itemize}
These results reflect the relative importance of multiscale components in the affine-control structure: fast dynamics dominate stability, while residual corrections fine-tune accuracy.

\paragraph{Progressive Deepening Until Saturation}
Further deepening the best-performing branches shows diminishing returns.
Performance improves clearly for the first additional layer but saturates after the second, with less than 1\% relative MSE gain.
This suggests that branch depth beyond this point no longer provides meaningful expressive advantage.

\begin{table}
    \centering
    \caption{Model sensitivity analysis results showing the marginal gain of expanding different branches.}
    \label{tab:sensitivity}
    \footnotesize
    \setlength{\tabcolsep}{4pt}
    \begin{tabular}{lcc}
        \toprule
        Model Variant             & MSE ($\times10^{-5}$) & Stability \\
        \midrule
        Baseline (Trunk-Base Opt) & 2.6853          & 0.8449 \\
        +1 Layer Base             & 2.6620          & 0.8489 \\
        +1 Layer Fast             & 2.8141          & \textbf{0.8518} \\
        +1 Layer Slow             & 2.8668          & 0.8432 \\
        +1 Layer Residual         & \textbf{2.5204} & 0.8492 \\
        +2 Layers Base            & 2.9092          & 0.8499 \\
        +2 Layers Residual        & 2.6606          & 0.8429 \\
        \bottomrule
    \end{tabular}
\end{table}

\subsection{\label{sec-exp-physical-consistency} Physical Consistency Evaluation}
This section evaluates whether the learned Neural CDE model preserves the key physically coherent behaviors discussed in Section~\ref{sec-theory}.

\paragraph{Oscillatory-Mode Preservation} Across all test operating conditions, as shown in Fig.~\ref{fig:current_dq} and Fig.~\ref{fig:current_abc}, the surrogate reproduces the dominant oscillatory mode of the grid current consistent with the EMT response, indicating that the learned model retains the principal damped-oscillation behavior present in the sampled data. We note that the high-frequency LC resonance itself is largely suppressed at the 1~kHz sampling rate, so this agreement reflects the effective closed-loop dynamics rather than the bare LC resonance.

\paragraph{Damping Consistency}
The damping score $S_{\text{damp}} = 0.85$ is computed across a range of test operating points. As shown in Fig.~\ref{fig:current_dq}, the learned model consistently reproduces the qualitative decay patterns of the grid current envelope following step perturbations. These results confirm that the surrogate captures the effective damping introduced by the current controller and modulation.

\paragraph{Long-horizon Stability.}
We assess the long-horizon stability of the learned Neural CDE model by performing extended rollouts over a duration significantly longer than the training trajectories. As shown in Fig.~\ref{fig:long_horizon}, the model maintains bounded state trajectories without exhibiting numerical divergence or unphysical growth. This behavior reflects the stability structure imposed by the inverter control loops in the EMT system, indicating that the learned continuous-time vector field remains well-behaved during extended integration.

\paragraph{Step-Size Reduction and Deployment Potential.}
The reference EMT trajectories are simulated with a $10~\mu$s step (100~kHz), whereas the learned surrogate reproduces the same dynamics while advancing at a $1$~ms step (1~kHz), corresponding to roughly two orders of magnitude fewer integration steps.
Since the adaptive inference solver can further enlarge the effective step where the dynamics are smooth, this step-size reduction indicates the potential to lower the computational burden in EMT-level studies.
Implementation-dependent wall-clock savings depend on solver coupling, hardware, and deployment details and will require further evaluation.

\section{Conclusion}

This work introduced a Neural CDE framework for modeling grid-forming inverter dynamics under discrete, multiscale control inputs.
The affine-control decomposition with dual fast/slow pathways enables the surrogate to capture both low-frequency supervisory behavior and high-frequency electromagnetic responses.
Ablation and sensitivity studies show consistent gains in accuracy and long-horizon stability, while physics-coherence evaluations confirm that the model preserves the dominant oscillatory modes, effective damping, and bounded trajectories.
Overall, Neural CDE-based component modeling provides a stable, accurate, and physically consistent surrogate suitable for EMT-level simulation studies and future large-scale deployment.

Looking forward, richer state measurements, such as full LC inductor currents and capacitor voltages, would enable enforcing energy-consistency constraints based directly on the physical LC energy function. Incorporating such energy-preserving priors into the Neural CDE training objective may further improve stability, enhance extrapolation across operating regimes, and yield surrogates that remain faithful to the underlying converter physics even under sparse supervision.

%Bibliography
\bibliographystyle{unsrtnat}
\bibliography{refs}

\end{document}